\documentclass[runningheads]{llncs}
\usepackage{graphicx}
\usepackage{multirow}
\usepackage{url}
\usepackage{todonotes}
\usepackage{amsmath,amssymb,amsfonts,bm}

\urldef{\mailsa}\path|{jimar,pkral,llenc}@kiv.zcu.cz, cerisara@loria.fr|

\usepackage{times}
\usepackage{latexsym}

\usepackage{graphicx}

\begin{document}

\title{Cross-lingual Approaches for Task-specific Dialogue Act Recognition}

%
%\titlerunning{Abbreviated paper title}
% If the paper title is too long for the running head, you can set
% an abbreviated paper title here
%

\author{Ji\v{r}\'i Mart\'inek$^{1,2}$, Christophe Cerisara$^{3}$, Pavel Kr\'al$^{1,2}$, Ladislav Lenc$^{1,2}$}

\institute{
        Dept. of Computer Science \& Engineering\\
        Faculty of Applied Sciences\\
        University of West Bohemia\\
        Plze\v{n}, Czech Republic\\
         \and
        NTIS - New Technologies for the Information Society\\
        Faculty of Applied Sciences\\
        University of West Bohemia\\
        Plze\v{n}, Czech Republic\\
         \and
        Universit\'e de Lorraine\\
        CNRS, LORIA, F-54000\\
        Nancy, France\\
\mailsa\\
}
\authorrunning{J. Martinek et al.}
\maketitle              % typeset the header of the contribution
\begin{abstract}
In this paper we exploit cross-lingual models to enable dialogue act recognition for specific tasks with a small number of annotations. We design a transfer learning approach for dialogue act recognition and validate it on two different target languages and domains. We compute dialogue turn embeddings with both a CNN and multi-head self-attention model and show that the best results are obtained by combining all sources of transferred information. We further demonstrate that the proposed methods significantly outperform related cross-lingual DA recognition approaches.

\end{abstract}

\section{Introduction}

Automatic dialogue act (DA) recognition has reached near-human performance on standard corpora, such as the Switchboard Dialogue Act corpus.
However, various application domains lead to different types of dialogues:
this variability, which is illustrated in the French corpus described in Section~\ref{sec:tcof}, severely impacts the direct application of a model trained
on a standard corpus to many other types of application tasks, as shown next.
Furthermore, developing DA-recognition models in other languages than English requires the costly annotation
of large-enough corpora.
We propose investigating cross-lingual transfer learning methods to reduce the amount of annotation
required for each new domain and language.

Transfer learning aims to reuse knowledge gained from a~large corpus to improve the models trained on a related task with few annotations.
We investigate in this work two sources of information from which we transfer knowledge: pre-trained English BERT sentence embeddings and an English corpus annotated with dialogue acts.
Two dialogue act recognition tasks are considered: appointment scheduling in German and casual conversations in French. The amount of French annotated dialogue acts is limited to a few hundred samples that may be annotated by one application developer within a few hours.
The definition of the dialogue acts are the same in English and German but different in French.

In addition to the relatively large resources available in English, we further assume the availability of an automatic translation system;
we will use for this purpose Google's translation.
Section~\ref{sec:model} describes our transfer learning strategy.

\section{Related Work}
% State-of-the-art results on well-known datasets
New approaches in the dialogue act recognition field are mainly evaluated on English datasets. Some methods have also been tested on other languages, such as Spanish (DIHANA  corpus~\cite{benedi2006design}), Czech \cite{Kral05a}, French  \cite{barahona2012building} and German (Verbmobil~\cite{jekat1995dialogue} corpus).

A nice review of the state-of-the-art of the domain is summarised in~\cite{ribeiro2019multilingual}, where 
the models typically reach 80\% of accuracy on standard DA datasets.
Nevertheless, the variety of DA labels in datasets is an obstacle for effective multi-lingual and multi-dataset research. Several interesting research efforts have thus emerged to define and exploit generic dialogue acts~\cite{ISO}.
However, in practice, the specific requirements of most target tasks
prevent a widespread usage of such standards.
We rather focus next on specific types of dialogues and task-related dialogue acts.

%transfer learning related work
Transfer learning has been quite a popular approach in deep learning in recent years. Such approaches have proven particularly useful in the computer vision and Natural Language Processing (NLP) domains. 
For instance, for automated pavement distress detection and classification \cite{gopalakrishnan2017deep} or for face verification \cite{cao2013practical}. A well-known transfer learning approach consists in using pre-trained word embeddings, such as Word2vec (W2V)~\cite{mikolov2013efficient}, 
ELMO~\cite{peters2018deep} and BERT~\cite{devlin2018bert}.

In standard transfer learning, information flows from the source to the target domain (one direction only). A related approach is multi-task learning~\cite{caruana1997multitask} where information flow across all tasks (usually more than two): information learned in each task may improve other tasks learning.

%BERT Related work in dacts
In the dialogue act recognition domain, Dai et al.~\cite{dai2020local} fine-tune BERT to classify a single utterance with quite good results, while Wu et al.~\cite{wu2020tod} propose task-oriented dialogue BERT (ToD BERT).

\section{Models}
\label{sec:model}
\subsection{English DA Classifier}

Our initial English dialogue act recognition model trained on the English dataset annotated
with dialogue acts is a multi-layer perceptron (MLP) with BERT embeddings as inputs.
We have tested various topologies for this initial model and chosen this one because of its
good performances and fast training times.

Each speaker turn, composed of a variable number of words, is first encoded into a single pre-trained 1024-dimensional sentence embedding vector with BERT Large\footnote{from https://github.com/google-research/bert\#pre-trained-models -- BERT-Large, Uncased (Whole Word Masking): 24-layer, 1024-hidden, 16-heads, 340M parameters}.

\begin{figure}[!ht]
\centering
\includegraphics[width=0.41\textwidth,angle=-90]{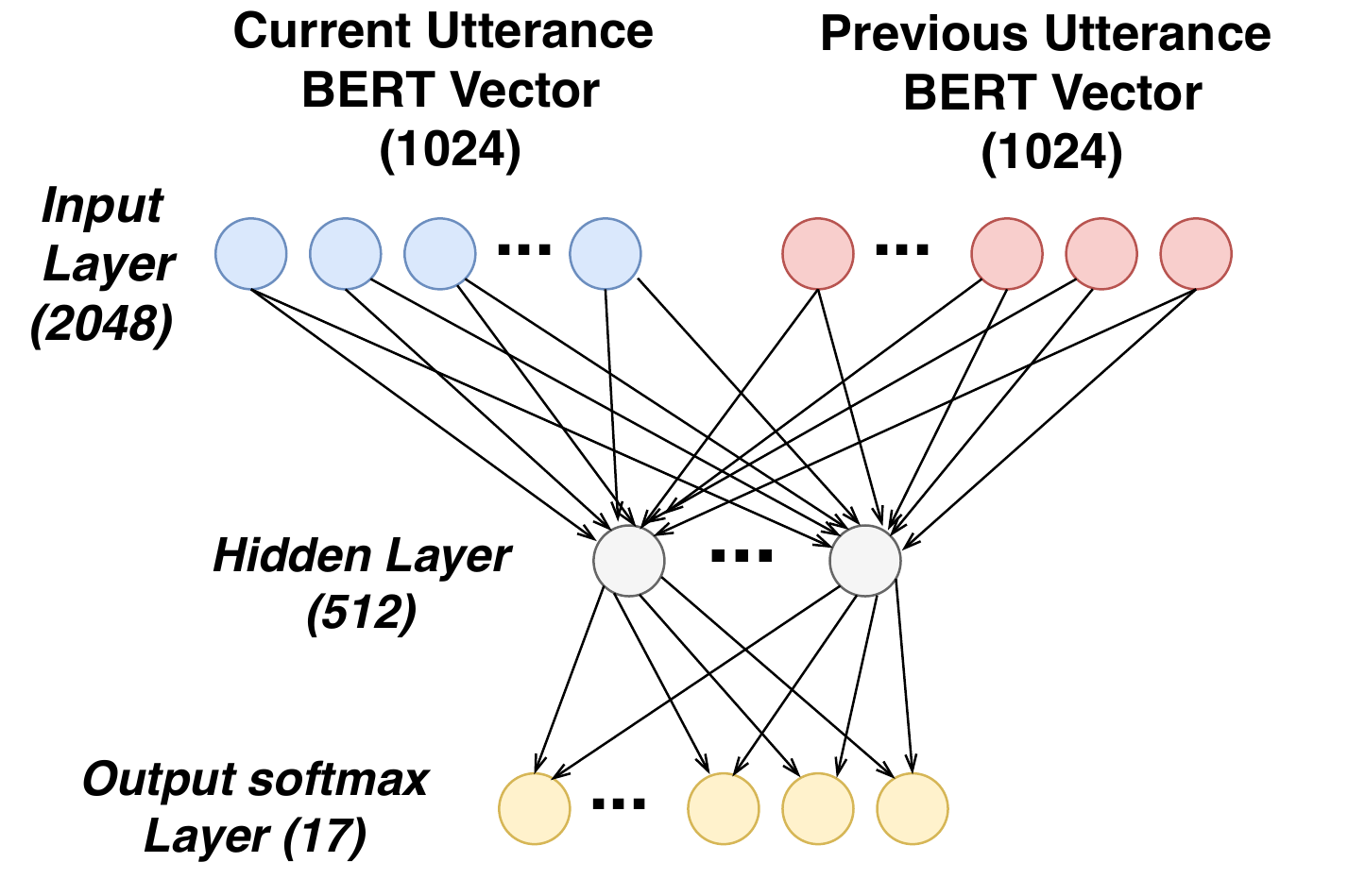}
\caption{English MLP model}
\label{fig:bert_mlp_model}
\end{figure}

As shown in Figure~\ref{fig:bert_mlp_model}, two such vectors are computed, respectively, for the previous, and the current speaker turns and concatenated as an input to the MLP.
The MLP outputs 17 tags, which correspond to the dialogue act labels of the Verbmobil-EN corpus.
This MLP has been trained on the training part of the Verbmobil-EN corpus.

\subsection{Speaker Turn Embeddings}

In our MLP model described previously, variable-length sentences are encoded into a unique speaker turn embedding vector with a pre-trained BERT model. We have further used two other models to compute the speaker turn embeddings: a convolutional neural network (CNN) and a multi-head self-attention (MH-SAtt) model.

\begin{figure}[!ht]
\centering
\includegraphics[width=0.72\textwidth,angle=-90]{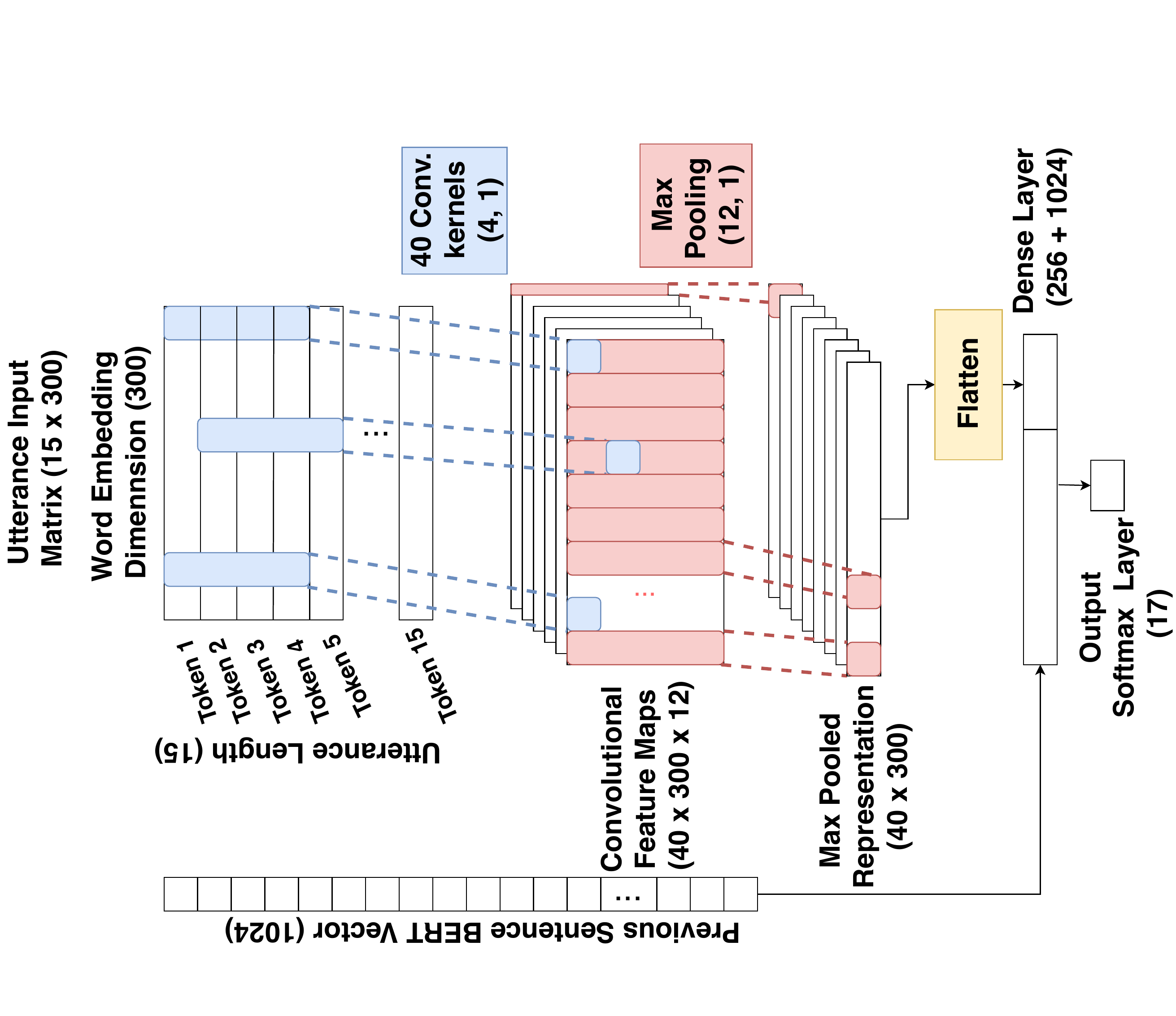}
\caption{CNN model for DA recognition %\cite{martinek2019multi}
}
\label{fig:cnn_model_for_dact}
\end{figure}

The CNN model, shown in Figure~\ref{fig:cnn_model_for_dact},
is derived from the model of Martinek et al.~\cite{martinek2019multi}.
It takes as an input a word sequence truncated/padded
to 15 words that always include the last two words of the sequence, following~\cite{cerisara17}.
These 15 words are encoded into either random embeddings or W2V vectors.
The CNN outputs a 256-dimensional vector for the current speaker turn.

\begin{figure}[!ht]
\centering
\includegraphics[width=0.75\textwidth,angle=0]{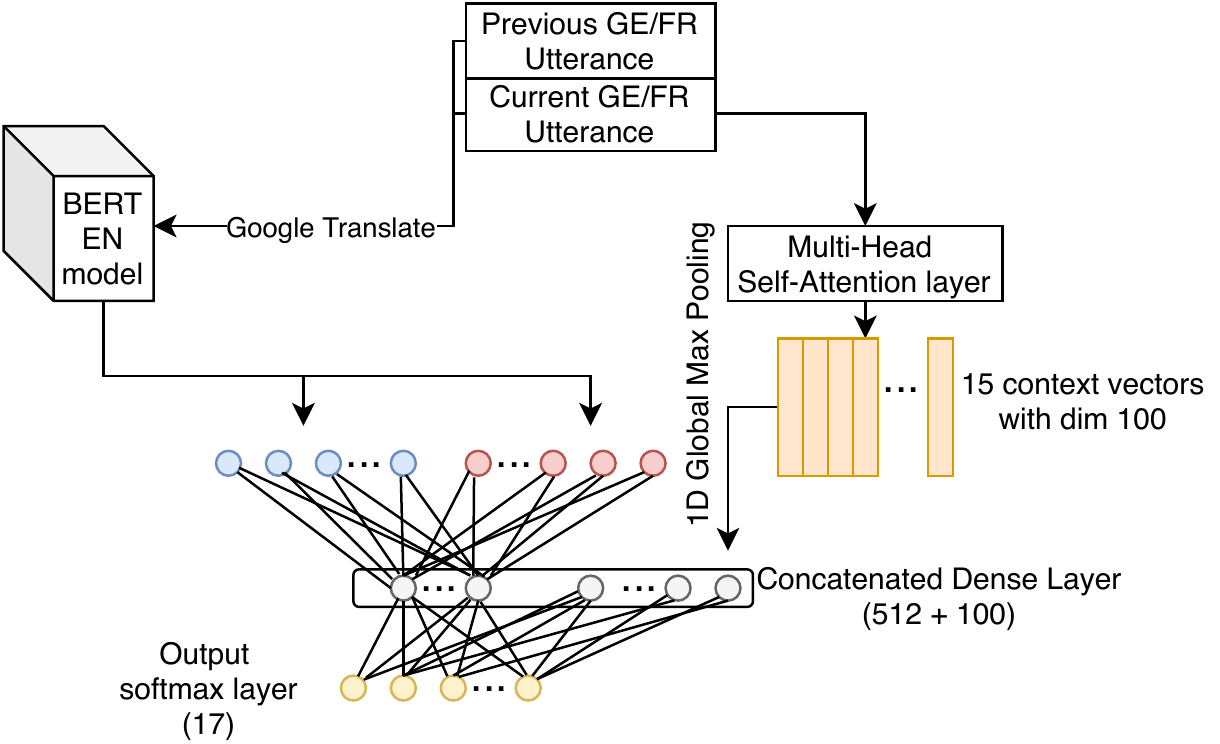}
\caption{Multi-head self-attention model}
\label{fig:selfatt}
\end{figure}

The MH-SAtt model transforms each input random word embedding with the standard scaled dot-product multi-head self-attention module~\cite{att}\footnote{https://github.com/CyberZHG/keras-multi-head}. A global max pooling operation is then applied to compute the speaker turn embedding.
Figure~\ref{fig:selfatt} shows how this model is used in our experiments.

In all our models, the previous speaker turn is also encoded into a 1024-dimensional vector with BERT and injected into the classification step:
this is an easy way to take into account the previous dialogue act without requiring a costly RNN, CRF and/or beam search process.

\section{Transfer Learning Approach}
Pre-trained BERT is famous for transferring general English lexical information into a large variety of downstream NLP tasks.
We exploit them for cross-lingual dialogue act recognition through automatic language translation. Similarly, we evaluate pre-trained word2vec English vectors in this context.
We further investigate the option to stack the translated inputs with the original foreign representations in the MH-SAtt model to increase the robustness of the model to translation errors.
Then, given a target non-English task with a few hundred annotated
samples only, we evaluate the benefit from fine-tuning the English DA classifier on this target small dataset.
Our global transfer learning approach thus consists of two phases:

\begin{enumerate}
    \item \textbf{Initial phase}\\
    The original English model in Figure~\ref{fig:bert_mlp_model} is trained on a large corpus annotated with dialogue acts.
    
    \item \textbf{Fine-tuning phase}\\
    \begin{itemize}
        \item The foreign sentences of the task-specific corpus annotated with dialogues acts are automatically translated into English with Google Translate\footnote{Any other translation system may also be used};
        \item The final classification layer (in Figures~\ref{fig:bert_mlp_model}, \ref{fig:cnn_model_for_dact}, \ref{fig:selfatt}) is replaced
        by a random layer with the same number of outputs than DAs
        in the target task;
        \item The model parameters are trained for a few epochs on the small target training corpus translated into English, as well as on the original foreign corpus for the MH-Satt model.
    \end{itemize}

\end{enumerate}

\begin{figure}[htb!]
\centering
\includegraphics[width=0.65\textwidth]{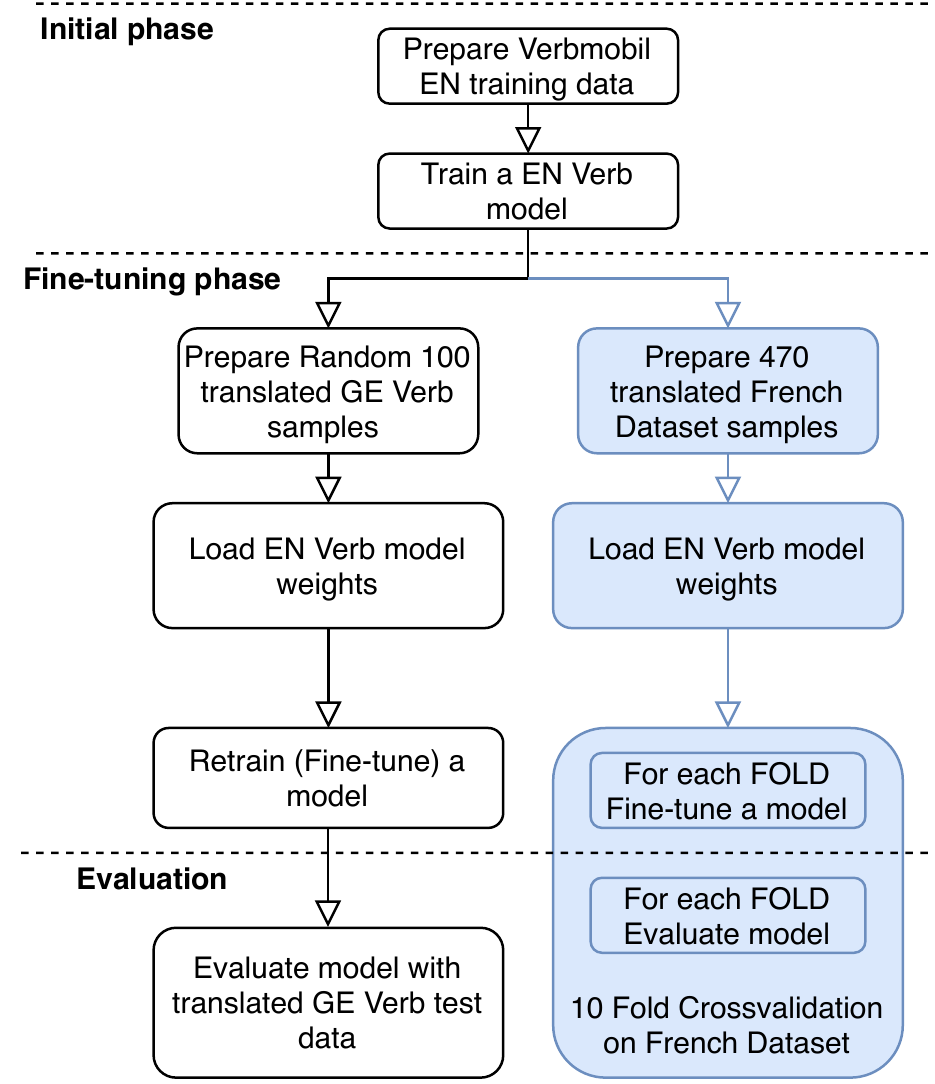}
\caption{Transfer learning process for both our experimental target tasks}
\label{fig:fine_tune_transfer_learning}
\end{figure}

The diagram shown in Figure~\ref{fig:fine_tune_transfer_learning} summarises the two phases for both target tasks and languages that we have used in our experiments.

%Fine-tuning the English model may be complemented by another type of transfer learning, inspired by the {\it stacked generalization}~\cite{stacking} ensemble method.
%It consists in stacking the English model's predictions as new (fixed) inputs into the small foreign classifier. 
%The advantage of this approach is that both models may have different architectures and also that it combines
%both foreign and translated inputs, increasing its robustness to potential translation errors.

\section{Experiments}

In the following experiments, we use two target tasks, one in German and another one in French.
Every experiment is run ten times and the results are averaged. The standard deviation is also computed.

\subsection{German Task}
Our first dialogue task consists in scheduling an appointment in German. Our target corpus is built by randomly sampling 100 utterances from the Verbmobil-GE training corpus
with the same dialogue acts distribution as in the complete training corpus, see Table~\ref{tab:trainingdistribution}. % \TODO{Jirka: show the training corpus distribution in Tab1}.
We could have used the full Verbmobil corpus for training, but this
would not have represented a realistic situation, where annotated 
training data for a specific task is usually scarce.
However, to better evaluate the quality of our approach, the model trained on this small corpus is tested on the complete (1460 utterances), standard Verbmobil-GE test corpus.

\begin{table}[htb!]
\centering
%{\scriptsize
\caption{Distribution of DAs in Verbmobil-GE corpus}
\begin{tabular}{|l|l|l|l|l|}
\cline{1-2} \cline{4-5}
\textbf{Label} & \textbf{Occurrence} & \textbf{} & \textbf{Label} & \textbf{Occurrence} \\ \cline{1-2} \cline{4-5} 
FEEDBACK       & 28\%               &           & DELIBERATE     & 3\%                \\ %\cline{1-2} \cline{4-5} 
SUGGEST        & 19\%               &           & INTRODUCE      & 2\%                \\ %\cline{1-2} \cline{4-5} 
INFORM         & 18\%               &           & COMMIT         & 1\%                \\ %\cline{1-2} \cline{4-5} 
REQUEST        & 9\%                &           & CLOSE          & 1\%                \\ %\cline{1-2} \cline{4-5} 
GREET          & 4\%                &           & POLIT. FORM.   & 1\%                \\ %\cline{1-2} \cline{4-5} 
BYE            & 4\%                &           & THANK          & 1\%                \\ %\cline{1-2} \cline{4-5} 
INIT           & 4\%                &           & DEFER          & 1\%                \\ %\cline{1-2} \cline{4-5} 
BACKCHANNEL    & 3\%                &           & OFFER          & 1\%                \\ \cline{1-2} \cline{4-5} 
\end{tabular}
\label{tab:trainingdistribution}
%}
\end{table}
%\vspace{-3pt}

\subsection{French Task}\label{sec:tcof}
The French tasks consist in manual transcriptions of spoken dialogues recorded by linguists in real-life situations involving volunteer citizens.
We manually annotate 470 turns from the TCOF corpus~\cite{tcof} with dialogue acts.
The types of dialogue in our 470 turns are very different from the ones found in standard DA corpora. They involve two friends trying to find a suitable gift, three students talking about their courses while chatting with someone else on their smartphone, and an adult talking with a young girl who is drawing.
The dialogue acts distribution is shown in Table~\ref{tab:orfeo}.
This French corpus is freely distributed with a CC BY-NC-SA license\footnote{https://github.com/cerisara/TCOFDA}.

\begin{table}[htb!]
\centering
%{\scriptsize
\caption{Distribution of DAs in the French corpus}
\begin{tabular}{|l|l|l|l|l|}
\cline{1-2} \cline{4-5}
\textbf{Label} & \textbf{Occurrence} & \textbf{} & \textbf{Label} & \textbf{Occurrence} \\ \cline{1-2} \cline{4-5} 
INFORM & 26\% & & OPEN ANSWER & 7\% \\
%\cline{1-2} \cline{4-5} 
AGREE & 13\% & & DISAGREE & 5\% \\
%\cline{1-2} \cline{4-5} 
BACKCHANNEL & 10\% & & YES ANSWER & 5\% \\
%\cline{1-2} \cline{4-5} 
Y/N QUESTION & 10\% & & NO ANSWER & 5\% \\
%\cline{1-2} \cline{4-5} 
OPEN QUESTION & 9\% & & OTHER ANSWERS & 1\% \\
%\cline{1-2} \cline{4-5} 
PERFORMATIVE & 8\% & & GREETINGS & $<1$\% \\
\cline{1-2} \cline{4-5} 
\end{tabular}
\label{tab:orfeo}
%}
\end{table}
%\TODO{pav: Table - open question != why question, however this is why question in the corpus!}

\subsection{Initial Phase: English Model}
The initial English MLP model is trained on the full Verbmobil-EN corpus.
The hyper-parameters of this model are tuned on the English corpus: this model is thus trained for 200 epochs with a learning rate of 0.002.
Table~\ref{tab:initial_phase_english} compares its accuracy when trained either from BERT embeddings
or from speaker turn embeddings obtained with the CNN model. The CNN may use either initial random or pre-trained W2V word embeddings. Every experiment is run ten times and the results are averaged.

\begin{table}[!htb]
%{\scriptsize
\caption{Initial Phase -- English MLP Model trained on Verbmobil-EN (9599 turns) and tested on Verbmobil-EN (1460 turns). The \textbf{Embeddings} column indicates how the word embeddings are initialised.}
\label{tab:initial_phase_english}
\begin{center}
\begin{tabular}{llccc}
\hline  \textbf{Model} & \textbf{Embeddings} & \textbf{Epochs} & \textbf{Test Acc} & \textbf{Std. Dev.} \\
\hline
\hline
MLP & BERT & 200 & 0.734 & 0.010 \\
CNN & W2V  & 200 & 0.704 & 0.009 \\
CNN & Random & 200 & 0.702 & 0.008 \\
\hline

\end{tabular}
\end{center}
%}
\end{table}

\subsection{Fine-Tuning Phase}

Fine-tuning consists in training the last classification layers of the models presented in Section~\ref{sec:model}
on the small foreign corpus.
This fine-tuning process thus involves some additional hyper-parameters, such as the learning rate and the fixed number of epochs,
which are tuned on a German development corpus composed of 10,000 utterances randomly sampled from the Verbmobil-GE training corpus. The same hyperparameter values found on this German corpus are also used for the French model\footnote{All hyperparameter values along with the source code are distributed with an open-source license in https://github.com/cerisara/TCOFDA}.

\subsection{Baseline Approaches}
Three baseline classifiers are shown in the first part of Table~\ref{tab:german_baseline}:
\begin{enumerate} %\itemsep {-1}
    \item \textbf{Majority Class Classifier (MC)}: it always predicts the most common DA in the training dataset.
    \item \textbf{Training from Scratch}:
Instead of transferring the model parameters from the initial English MLP model, the parameters (without the embeddings) are randomly initialised in this baseline.
    \item \textbf{No-Fine-Tuning}: The model parameters are simply transferred from the initial English MLP model, and no further training is done.
\end{enumerate}

\subsection{Fine-Tuning Experiments}

\subsubsection{German Results}

Table~\ref{tab:german_baseline} shows the results of our models
on the Verbmobil-GE test set.
All these models process only translated English sentences, except for the last MH-SAtt model that further includes the original German words, as shown in Figure~\ref{fig:selfatt}.

\begin{table}[htb!]
%{\scriptsize
\caption{Accuracy on the Test Verbmobil-GE corpus}
\begin{center}
\begin{tabular}{llccc}
\hline {\bf ~Model} & \textbf{Embeddings} &  \textbf{Epochs} & \textbf{Test Acc} ~ & \textbf{Std. Dev.} \\
\hline
\hline
\textit{Baseline MC} &  & -- & 0.279 & -- \\
\hline
\hline
\textit{From scratch} & & & & \\
~~~~CNN & W2V  & 15 & 0.410 & 0.012  \\
~~~~MLP & BERT & 25 & 0.468 & 0.008  \\
\hline
\hline
\textit{No-Fine-tuning} &  & &  &\\
~~~~CNN & W2V & -- & 0.380 \\
%~~~~CNN & Random & -- & 0.385 \\ %pav-deleted after discussion wirk Jirka
~~~~MLP & BERT & -- & 0.479\\
\hline
\hline
\textit{Fine-Tuning} & & & & \\
~~~~CNN & W2V  & 15 & 0.463 & 0.008 \\
~~~~MLP & BERT & 25 & 0.484 & 0.025 \\
~~~~MH-SAtt & BERT + GE & 50 & 0.502 & 0.012 \\
\hline
\end{tabular}
\end{center}
\label{tab:german_baseline}
%}
\end{table}

With regard to transfer from pre-trained models, BERT is consistently better than W2V.
Simply reusing the English models without fine-tuning gives the worst results, while fine-tuning the English models on the small target corpus systematically improves the results. The best performances are obtained
with the MH-SAtt model that combines both BERT pre-trained vectors and
fine-tuning transfer learning with the original and translated words sequences.

\subsubsection{French Results}
We carried out 10-fold cross-validation on the 470 speaker turns of the French corpus. As we do not have enough data to create a development corpus in French, we share the same hyper-parameters as found on the German corpus.
Table~\ref{tab:french_baseline} shows the results of this experiment.

\begin{table}[htb!]
%\scriptsize
\caption{French cross-validation experiments}
\begin{center}
\begin{tabular}{llccc}
\hline {\bf ~Model}~& \textbf{Embeddings} & \textbf{Epochs} & \textbf{Acc} ~ & \textbf{Std. Dev.}\\
\hline
\textit{Baseline MC} & ~ & -- & 0.210 & -- \\
\hline
\hline
\textit{From scratch} & & & & \\
~~~~CNN & W2V & 15 &  0.403 & 0.006 \\
~~~~MLP & BERT & 25 &  0.421 & 0.005 \\
\hline
\hline
\textit{Fine-Tuning} & & & \\
~~~~CNN & W2V & 15 & 0.400 &  0.009 \\
~~~~MLP & BERT & 25 & 0.430 & 0.008 \\
~~~~MH-SAtt & BERT + FR & 50 & 0.436 & 0.007 \\
\hline
\end{tabular}
\end{center}
\label{tab:french_baseline}
%}
\end{table}

The relative contributions of the various sources of information and models are similar to the German experiments.
However, the differences between the results are much smaller, which is likely due to the fact that the hyper-parameters have not been tuned on
French but on German.

\subsection{Comparison with Related Work}
We compare in Table~\ref{tab:sota} our proposal with other cross-lingual methods for dialogue act recognition that we are aware of.
This related work projects German word embeddings into the English
word embeddings space with the canonical correlation analysis method (CCA)~\cite{brychcin2020linear}.
The comparison is done on the German Verbmobil test corpus.
Table~\ref{tab:sota} clearly shows that the proposed methods are significantly better than the previous work.
To the best of our knowledge, no other study dealing with cross-lingual dialogue act recognition has been published.

\begin{table}[htb!]
%{\footnotesize
\caption{Comparison with previous cross-lingual DA recognition on the German Verbmobil test corpus}
\begin{center}
\begin{tabular}{lc}
\hline {\bf ~Methods}~& {\bf Acc}\\
\hline
%Baseline MC & 0.210 \\
CNN + CCA~\cite{martinek2019multi} & 0.315 \\
BiLSTM + CCA~\cite{martinek2019multi} & 0.340 \\
\hline
MLP + BERT & 0.484 \\
MH-SAtt + BERT + GE & 0.502 \\
\hline
\end{tabular}
\end{center}
\label{tab:sota}
%}
\end{table}

\section{Conclusions}

We explore two types of transfer learning for 
cross-lingual DA recognition: pre-trained word embeddings and classifier fine-tuning. 
Three types of dialogue turn embeddings are computed, based respectively on BERT, CNN and multi-head self-attention.
The objective is to leverage large available English resources annotated in dialogue acts to enable DA recognition on a specific target task and language with a limited amount of annotations.

We have validated these approaches on two tasks and languages, German and French, and released a dedicated French DA corpus with real-life dialogues recorded in quite different conditions than the
existing standard DA corpora.
The best results are obtained when 
all available sources of information are included, with both
the BERT and multi-head self-attention dialogue turn embeddings.

We further demonstrated that the proposed methods significantly outperform cross-lingual DA recognition approaches developed previously.

\section*{Acknowledgements}
%ADD in CR version
This work has been partly supported by Cross-border Cooperation Program Czech Republic - Free State of Bavaria ETS Objective 2014-2020 (project no. 211), by Grant No. SGS-2019-018 Processing of heterogeneous data and its specialized applications and by GENCI-IDRIS (Grant 2021-AD011011668R1).

\bibliographystyle{splncs04}
\bibliography{paper}

\end{document}